\documentclass{article}
\usepackage[letterpaper]{geometry}
\usepackage{spconf,graphics}
\usepackage[bookmarks=false]{hyperref}

\usepackage{graphicx}
\usepackage{caption}
\usepackage{subcaption}
\usepackage{tabularx}
\usepackage{cite}                      % needed to automatically sort the references
\usepackage{enumitem}

\title{Accurate Human Body Reconstruction for Volumetric Video}

\name{Decai Chen, Markus Worchel, Ingo Feldmann, Oliver Schreer, Peter Eisert}
\address{Fraunhofer Institute for Telecommunications, Heinrich Hertz Institute, Berlin, Germany}

\begin{document}
\maketitle

\begin{abstract}
In this work, we enhance a professional end-to-end volumetric video production pipeline to achieve high-fidelity human body reconstruction using only passive cameras. While current volumetric video approaches estimate depth maps using traditional stereo matching techniques, we introduce and optimize deep learning-based multi-view stereo networks for depth map estimation in the context of professional volumetric video reconstruction. Furthermore, we propose a novel depth map post-processing approach including filtering and fusion, by taking into account photometric confidence, cross-view geometric consistency, foreground masks as well as camera viewing frustums. We show that our method can generate high levels of geometric detail for reconstructed human bodies.
\end{abstract}


%%%%%%%%% BODY TEXT
\section{Introduction}

\begin{figure}[tb]
 \centering % avoid the use of \begin{center}...\end{center} and use \centering instead (more compact)
 \includegraphics[width=\columnwidth]{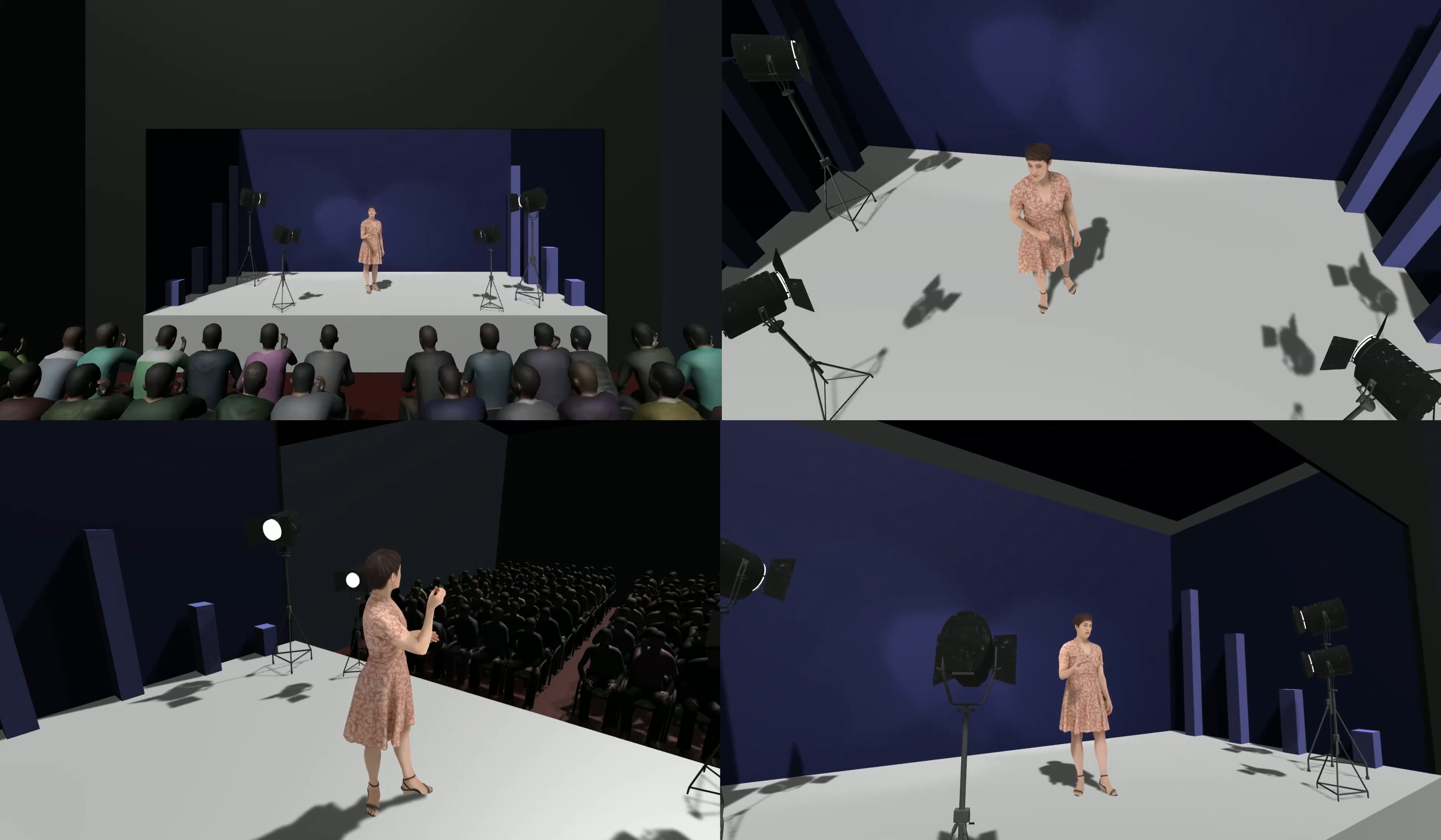}
 \caption{A VR application from our work uses a basic CG environment to underline the realistic and natural impression of real person based volumetric video assets.}
 \label{fig:arvr}
\end{figure}

With rapid development and widespread popularity of Augmented Reality (AR) and Virtual Reality (VR), Volumetric Video, also known as Free-Viewpoint Video, is playing a more and more important role as it bridges the gap between the real and virtual world. Volumetric video has a variety of applications including telecommunication, movie production, games, sports broadcasting, cultural heritage recovery, remote planning, as well as therapy and rehabilitation, as it provides immersive user experiences and realistic scenes from arbitrary unconstrained viewpoints. \autoref{fig:arvr} shows an example of a VR experience production using our volumetric reconstruction assets. We notice that high-quality reconstruction and representation of human performances is essential for applications of volumetric video.

% While volumetric video systems like \cite{collet2015high} show great potential for high-quality reconstruction, their sophisticated setups with active depth cameras are less desirable. Instead of reconstructing a general scene, recent progress in deep learning draws much attention to predict parametric human body models from images \cite{dibra2016hs,dibra2017human,kanazawa2018end,tan2017indirect,SMPL:2015,huang2020arch,zheng2019deephuman}. However, parametric approaches require precomputed models of all objects in the scene and thus cannot handle unseen scenarios such as diverse clothes, hairstyles and props due to lack of prior knowledge. In contrast, we revise a recent volumetric reconstruction pipeline \cite{worchel2020ernst}, to achieve accurate general 3D reconstruction of mainly but not limited to human body, at a fine geometry level using only RGB cameras and without prior knowledge of template models.

% two separate paragraphs before
While volumetric video systems like \cite{collet2015high} show great potential for high-quality reconstruction, their sophisticated setups with active depth cameras are less desirable. In contrast, we revise a recent volumetric reconstruction pipeline \cite{worchel2020ernst}, to achieve accurate general 3D reconstruction of mainly but not limited to human body, at a fine geometry level using only RGB cameras. In this work, we optimize and adapt a state-of-the-art multi-view stereo (MVS) network called Vis-MVSNet\cite{zhang2020visibilityaware}, to take the advantage of deep feature representation and visibility guided regularization, for accurate depth map estimation under the existing setup of multi-view cameras in a capture studio \cite{schreer2019capture}. Moreover, inspired by BlendedMVS \cite{yao2020blendedmvs}, we create a 3D human body dataset captured in a studio environment, to fine-tune the neural networks with the knowledge of studio images and given human bodies. 

% To capture the scene geometry from 360 degrees, it is desirable to distribute cameras in the space of a studio. The key is to estimate an accurate depth map for each camera and fuse them appropriately into a global space. Most of volumetric reconstruction frameworks \cite{collet2015high,schreer2019capture,worchel2020ernst,orts2016holoportation} group cameras into spatial pairs and estimate per-pair depth maps, using variants of PatchMatch Stereo \cite{bleyer2011patchmatch} for group-wise stereo matching. In contrast, from the algorithmic point of view, we break the limitation of camera pairs by multi-view stereo depth estimation, which integrates information from all neighboring views rather than the closest view. 

% \begin{figure}[tb]
%  \centering
% %  \includegraphics{horizontal_crop.jpg}
%  \includegraphics[width=\linewidth]{horizontal_crop.jpg}
%  \caption{Example for a VR production based on our volumetric reconstruction. It illustrates how real persons can be integrated as volumetric video assets into virtual environments. In this example, the reconstructed persons act interactively with each other in a virtual scene.}
%  \label{fig:rose_teaser}
% \end{figure}

% While traditional stereo matching \cite{bleyer2011patchmatch,Waizenegger2016} and multi-view stereo (MVS)\cite{galliani2015massively,schoenberger2016mvs,Xu2019} approaches typically calculate matching cost within local patches of raw images, they fail to represent context information including semantics, thus may rely on post global optimization like semi-global matching (SGM)\cite{Hirschmuller2008}. Recent convolutional neural networks (CNNs) such as Feature Pyramid Network (FPN)\cite{lin2017feature} and U-Net \cite{ronneberger2015u} show significantly better performance on patch matching \cite{JMLR:v17:15-535}.
% In this work, we optimize and adapt a state-of-the-art multi-view stereo (MVS) network called Vis-MVSNet\cite{zhang2020visibilityaware}, to take the advantage of deep feature representation and visibility guided regularization, for accurate depth map estimation under the existing setup of multi-view cameras in a capture studio \cite{schreer2019capture}. Moreover, inspired by BlendedMVS \cite{yao2020blendedmvs}, we create a 3D human body dataset (HBR dataset) captured in a studio environment, to fine-tune the neural networks with the knowledge of studio images and given human bodies. 

In addition, we propose a novel post-processing approach for filtering and fusing depth maps, to obtain high-quality point clouds for mesh reconstruction. Different from standard depth map filtering techniques in MVS \cite{yao2018mvsnet, zhang2020visibilityaware,yan2020dense,schoenberger2016mvs,Xu2019,gu2020cascade} which only remove points either geometrically inconsistent or with low photometric confidence, we further filter out points falling in the background mask of any camera and outside the visual frustum of a certain number of cameras \cite{worchel2020ernst}.

% In sum, the main contributions of this work are:
% \begin{itemize}
%   \item A novel pipeline for highly accurate reconstruction of volumetric video is presented with enhanced deep learning-based depth estimation.
%   \item We propose a new post-processing method for robust and effective depth map filtering and fusion.
%   \item We demonstrate significantly enhanced quality of our work compared to existing approaches.
% \end{itemize}
In summary, the main contributions of this work are: (\romannumeral 1) A novel pipeline for highly accurate reconstruction of volumetric video is presented with enhanced deep learning-based depth estimation. (\romannumeral 2) We propose a new post-processing method for robust and effective depth map filtering and fusion. (\romannumeral 3) We demonstrate significantly enhanced quality of our work compared to existing approaches via extensive experiments.

\section{Related Work}
\textbf{Volumetric Capture System.} 
% \cite{4DviewsV32,8iThird87,MixedRea34,HomeVolu96,HomeMet84}
A multi-camera capture system is essential for end-to-end dynamic 3D reconstruction solutions, where human bodies usually serve as the reconstruction subject.
This technique has been available for commercial use in recent years \cite{MixedRea34,HomeVolu96,HomeMet84}, meanwhile drawing more and more attention in academic research. Collet et al.\cite{collet2015high} build a capture system with 106 cameras including RGB and infrared (IR) sensors, as well as IR illuminators, and combine multiple modality for high-quality geometry reconstruction, specifically RGB stereo, active IR stereo, and Shape from Silhouette. Similarly, Orts-Escolano et al. \cite{orts2016holoportation} utilize 8 trinocular pods, each consisting of 1 RGB and 2 IR cameras, as well as an IR projector, to achieve real-time 3D reconstruction. Huang et al. \cite{huang2018deep} reconstruct human performance with a sparse passive capture system using a learning-based volumetric approach, but only to a low level of geometric accuracy. More recently, Guo et al. \cite{Relightables} present a promising photorealistic and relightable human reconstruction pipeline by sophisticated hardware setup. In addition to 16 IR structured light projectors, 58 RGB and 32 IR cameras, it requires 331 programmable LEDs to provide alternating lighting patterns at 180 Hz. In contrast, our method relies on only 32 RGB cameras without any active sensors and thus is more accessible.

\textbf{Multi-view Stereo.}
% Given two or multiple calibrated images, stereo matching and MVS restore 3D geometry of the scene. While the 3D geometry in stereo matching is typically represented by disparity maps (for rectified input image pairs) or sometimes depth maps, MVS has four types of representations: direct point clouds, volume grids, mesh surfaces and per-view depth maps. Due to flexibility and conciseness, depth map based MVS gains the most popularity.
Traditional stereo matching and MVS methods compute pairwise matching cost of raw image patches by Sum of Absolute Differences (SAD), Sum of Squared Distances (SSD) or Normalized Cross-Correlation (NCC). In recent years, PatchMatch-based stereo matching \cite{bleyer2011patchmatch} and MVS \cite{galliani2015massively,schoenberger2016mvs,Xu2019} are dominating traditional methods due to highly parallelism and robust performance. Note that most of volumetric capture pipelines employ either traditional PatchMatch-based stereo matching \cite{collet2015high,orts2016holoportation} or MVS \cite{Relightables}. Recently, deep learning shows superior performance in MVS. MVSNet \cite{yao2018mvsnet} applies a similar idea to stereo matching, by regularizing cost volumes with 3D CNNs. The main difference is that MVS builds cost volumes by warping feature maps from multiple neighboring views instead of one. To reduce memory consumption from 3D CNNs, R-MVSNet \cite{yao2019recurrent} sequentially regularizes 2D cost maps with a gated recurrent network (GRU), while other works \cite{gu2020cascade,cheng2020deep,yang2020cost} integrate multi-stage coarse-to-fine strategies to progressively refine 3D cost volumes. Moreover, Vis-MVSNet \cite{zhang2020visibilityaware} explicitly estimates pixel-wise visibility as certainty to guide multi-view cost volume fusion, leading to more robust performance. More recently, Wang et al. \cite{wang2020patchmatchnet} integrate an iterative multiscale PatchMatch into a trainable MVS architecture. After exploring a variety of state-of-the-art MVS networks, we find that Vis-MVSNet performs best in the field of volumetric reconstruction, and we further optimize and adapt it into our volumtric human reconstruction pipeline.
% However, traditional methods fail to represent semantic context information, thus may rely on post global optimization like semi-global matching (SGM)\cite{Hirschmuller2008}.

% Recently, deep learning shows superior performance in both stereo matching and MVS. In terms of the former, PSMNet \cite{chang2018pyramid} introduces spatial pyramid pooling (SPP) and utilizes a 3D hour-glass network for cost volume regularization. More recently, AANet \cite{xu2020aanet} proposes sparse points based intra-scale cost aggregation layers with deformable convolution, together with cross-scale cost aggregation layers. When it comes to learning-based MVS,

\begin{figure*}[t]
 \centering 
 \includegraphics[width=\textwidth]{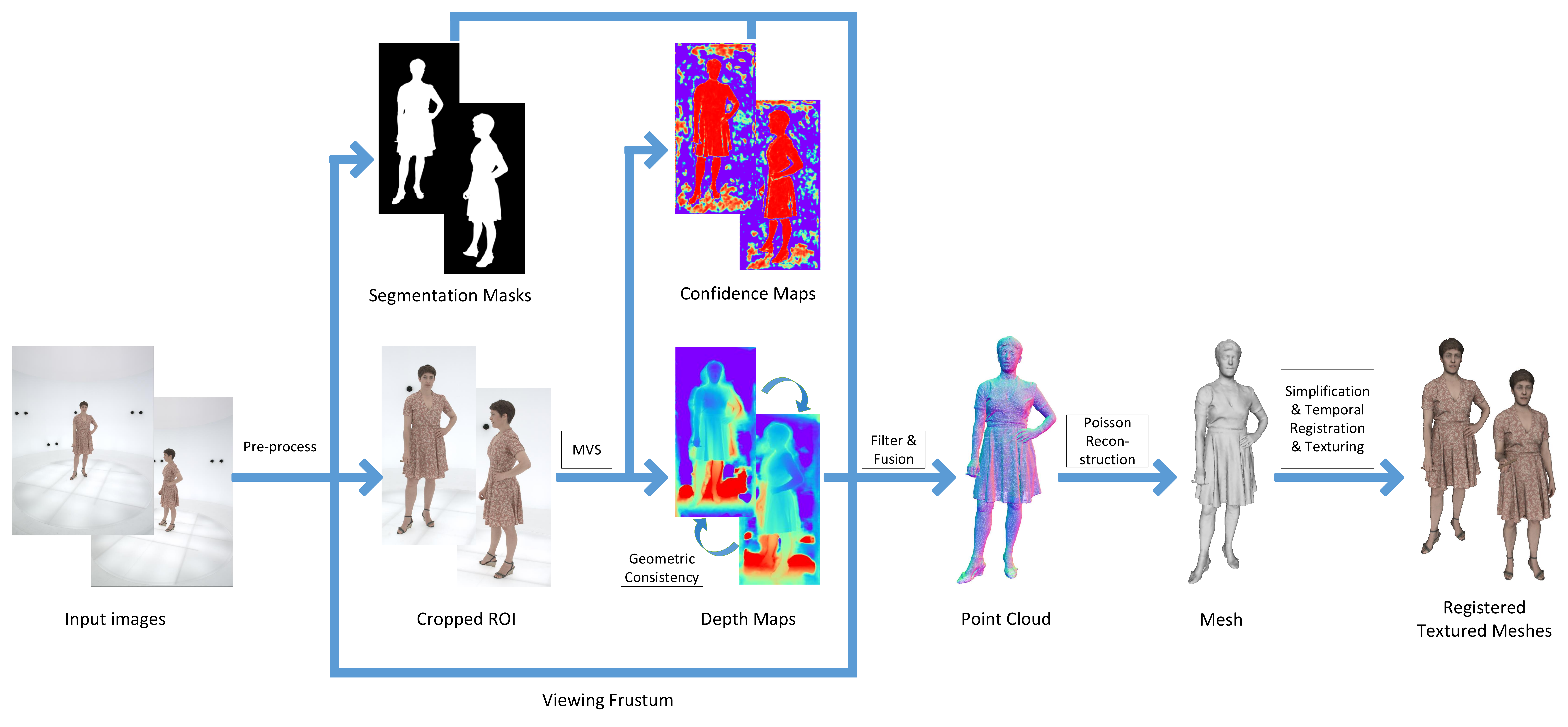}
 \caption{Overview of our volumetric reconstruction pipeline. Given 32 input images per frame, we first crop the ROIs before estimating per-view depth maps using deep MVS networks. Subsequent depth map filtering and fusion is applied by taking into account MVS confidence maps, geometric consistency, foreground segmentation masks and viewing frustum to get robust geometry. Finally, meshes are reconstructed and temporally registered before encoded into a streamable MP4 file.}
 \label{fig:pipeline}
\end{figure*}

\textbf{Depth Map Fusion}
% Voxel-based fusion and point-based fusion are the two most common ways for fusing per-view information such as depth maps into common geometry. Curless et al. \cite{curless1996volumetric} introduce Voxel-based fusion with discretized regular grid to store signed distance functions (SDF) that implicitly represent the underlying geometry. To improve the memory efficiency of voxel-based representations, Steinbrücker et al. \cite{steinbrucker2013large} hierarchically represent the scene by multi-scale octrees, while Nießner et al. \cite{niessner2013real} use spatial hashing functions to compress the space, enabling real-time data access and update. Alternatively, depth maps can also be directly accumulated into a point cloud. Note that depth map filtering is usually performed before fusion, to discard unreliable or unconsistent points. Specifically, Galliani et al. \cite{galliani2015massively} propose geometrical consistency checks with respect to both depth and normal information. Sch\"{o}nberger et al. \cite{schoenberger2016mvs} cluster consistent pixels that are combined into the reconstructed point cloud, and discard clusters that are not supported by a certain number of views. As recent learning-based MVS approaches predict per-view depth maps together with confidence maps (also known as probability maps) \cite{yao2018mvsnet,yao2019recurrent,gu2020cascade,zhang2020visibilityaware}, the confidence map can serve as another criterion for depth map filtering. 
Voxel-based fusion and point-based fusion are the two most common ways for fusing per-view information such as depth maps into common geometry. Both categories of depth map fusion exist in volumetric reconstruction. Point-based pipelines \cite{schreer2019capture,collet2015high,Relightables} fuse refined depth maps into a point cloud before applying Screened Poisson surface reconstruction \cite{kazhdan2013screened} for meshing. In contrast, Voxel-based pipelines fuse depth maps into a Truncated Signed Distance Function (TSDF) volume \cite{curless1996volumetric}. Given this implicit volumetric representation, Orts-Escolano et al. \cite{orts2016holoportation} directly extract triangle meshes using the Marching Cubes level set method \cite{lorensen1987marching}, while Worchel et al. \cite{worchel2020ernst} first extract an oriented point cloud from the iso-surface and then perform Screened Poisson surface reconstruction for meshing. Inspired by the latter approach, we propose a novel depth map post-processing strategy consisting of filtering and fusion, by leveraging a large number of filtering criteria to remove unreliable estimations and the voxel-based fusion to suppress noise and outliers, achieving higher accuracy of the fused geometry.

\section{System Overview} \label{Overview}
% full 360 degree capturing \cite{schreer2019capture}.
In this section, we present an overview of our volumetric reconstruction pipeline. The capture system consists of 32 RGB cameras spatially arranged in 16 stereo pairs and distributed around a cylinder of 6m diameter and 4m height, in order for full 360 degree capturing. Unlike other volumetric capture systems \cite{collet2015high,orts2016holoportation,Relightables}, no active sensors such as IR illuminators and IR cameras are required. Instead of green screen background and directed light, diffuse lighting from omni-directional LED panels and a white background enables simple keying and better relighting for AR/VR appearance. Before volumetric performance capturing, all cameras are geometrically calibrated and photometrically matched to a reference camera, providing consistent cross-view appearance.

\autoref{fig:pipeline} demonstrates the overview of our pipeline. The inputs are 32 calibrated images of 20 MPixel resolution capturing at 25 Hz. A statistical background model for each camera is computed from clean plate recordings, simplifying foreground background segmentation. In order to capture dynamic and flexible human performances with possibly large movements or multiple humans using limited amount of cameras, each camera covers the whole space of potential human activities. However, this leads to the fact that only a small portion of the image is occupied by the human, while the major part filled by background is useless for reconstruction. Therefore, processing the whole high-resolution images does not only waste much computational resources and time, but also poses difficulty for performing inference or training of modern CNNs based algorithms, which typically demand considerable GPU memory. Specifically, we find that most of the learning-based state-of-the-art depth estimation networks including stereo matching \cite{chang2018pyramid,xu2020aanet} and MVS \cite{yao2018mvsnet,yao2019recurrent,gu2020cascade,zhang2020visibilityaware} do not support our full 20 MPixel resolution input images even on a graphic card of 24 GB memory, due to the memory-demanding nature of CNNs. To handle high-resolution images, we crop a region of interest (ROI) from the original image and only use the ROI in the following workflows, as shown in \autoref{fig:pipeline}. This can be implemented by applying a bounding box on the segmented foreground mask. Camera parameters are adjusted accordingly.

Given 32 images per frame, we use an optimized and tuned Vis-MVSNet \cite{zhang2020visibilityaware} for MVS depth estimation, which is explained in detail in \autoref{DepthMapEstimation}. After that, a novel depth map post-processing method is used to get a smooth but detailed point cloud, discussed in \autoref{DepthMapFusion}. Triangle meshes are extracted by Screened Poisson Surface Reconstruction \cite{kazhdan2013screened} and then simplified to the desired level of detail. Finally, the dynamic mesh sequence is temporally registered \cite{morgenstern2019progressive} as well as textured, before being compressed and encoded into a streamable MP4 file.

\section{Depth Map Estimation} \label{DepthMapEstimation}
The success of CNNs in feature extraction and cost volume regularization provides a promising way for estimating depth on areas which are challenging for classical methods, such as textureless, reflective, or occluded surfaces. After comparing state-of-the-art MVS and stereo matching approaches for depth estimation of volumetric reconstruction (see \autoref{Ex_DE} for details), we adapt an optimized and tuned Vis-MVSNet into our volumetric reconstruction pipeline. Given one reference image and some neighboring source images with calibration parameters, Vis-MVSNet estimates a depth map for the reference image, by improving the coarse-to-fine three-stage cascade MVSNet \cite{gu2020cascade} with consideration of pixel-wise visibility (or occlusion). 
% In this section, we first review the standard Vis-MVSNet \cite{zhang2020visibilityaware} before we present our improvement and adaption.

% First of all, a 2D U-Net \cite{ronneberger2015u} is applied to the input images for multi-scale feature extraction. The extracted features at the last three scales in the decoder part are used to construct three-staged hierarchical cost volumes. In each stage, each pair-wise reference-source cost volume is constructed by stacking cost maps for all depth hypotheses, which are calculated from homography warping followed by group-wise correlation \cite{guo2019group}. Each cost volume is regularized via 3D CNNs and produces an uncertainty map via soft-argmax and entropy operation. The uncertainty maps from each reference-source pairs serve as the weighting guidance for fusing latent volumes, and the fused volume is further regularized to regress the final depth map and probability map of the current stage, with the same resolution to the input feature map. Intermediate depth maps from previous stages are used for the cost volume construction at the next stage. The depth map from the last stage serves as the final output. Due to the cascaded coarse-to-fine architecture, Vis-MVSNet alleviates the expensive memory requirement of 3D CNNs, thus suitable for high-resolution reconstruction. Moreover, explicitly integrating the pixel-wise visibility by estimating matching uncertainty enables robust reconstruction even for occluded areas.

Defining an appropriate depth search range and sampling interval plays an important role in cost volume based depth estimation. A larger search range covers a bigger space for reconstruction while a smaller sampling interval tends to restore finer spatial details, but they require a larger number of depth samples thus increasing the computational overhead. Rather than setting a uniform depth range for all cameras, we assume a global cylinder inside which the captured human performance is located and project the cylinder boundary to each camera, to get the maximum and minimum depth value as its adaptive depth search range. 

Instead of performing uniform sampling in depth domain like many other approaches \cite{gu2020cascade,zhang2020visibilityaware,yao2018mvsnet}, we distribute sampling hypotheses uniformly along the epipolar line in inverse depth space (in other words, in disparity space):
\begin{equation}
d_{i}=\left(\frac{1}{d_{\max }}+\left(\frac{1}{d_{\min }}-\frac{1}{d_{\max }}\right) \frac{i}{D-1}\right)^{-1},
\end{equation}
where $i\in\{0,1,...,D-1\}$ is the index of a sampling hypothesis, $[d_{min},d_{max]}$ is the adaptive depth range for the reference image and $D$ is the total amount of sampling hypotheses. This sampling strategy builds a more discriminative cost volume and thus is more robust for large-scale scenes \cite{xu2020learning}. Different from the standard Vis-MVSNet, which shrinks the depth search range between consecutive stages by a scaling factor of 1/4, we instead apply a more aggressive factor of 1/8, which results in significantly finer resolution of depth sampling. The intuition behind this improvement is that in general complex scenes there exist a lot of occlusions between foreground and background objects and thus suffer from more depth discontinuities. Due to depth map up-sampling from coarse to fine stage, the interpolated depth values near depth discontinuities may go beyond the depth search range if this range is narrowed down aggressively, while a human body is generally a continuous surface with relatively less self-occlusions and therefore benefits from fast shrinkage of the depth search range. A comparison between the original Vis-MVSNet and our optimized one is shown in \autoref{Ex_DE}.

\begin{figure}[tb]
 \centering % avoid the use of \begin{center}...\end{center} and use \centering instead (more compact)
 \includegraphics[width=\columnwidth]{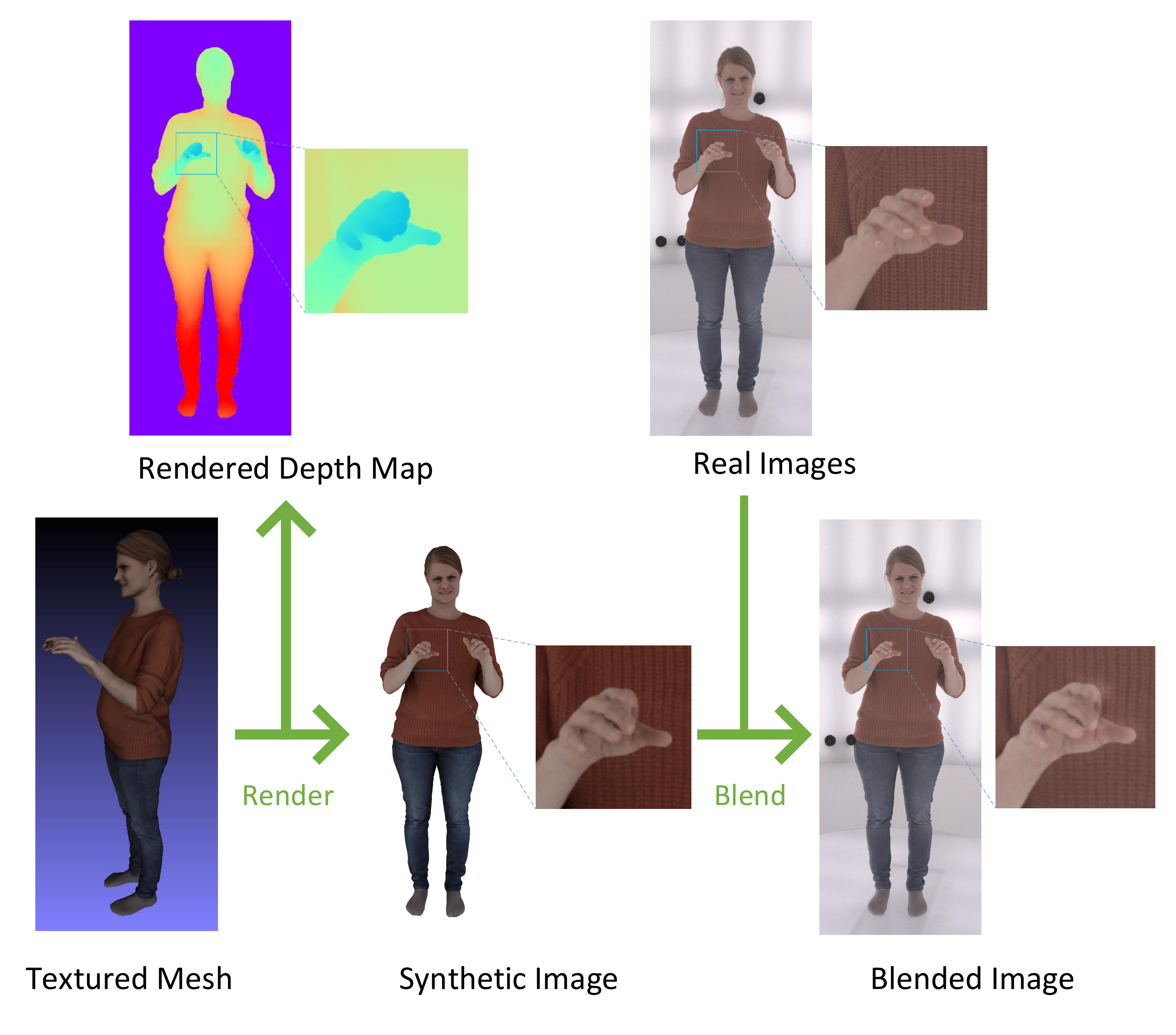}
 \caption{Generation of training data. Given a reconstructed textured 3D model, we render synthetic images and depth maps. Then, low frequency information from the real image and the high frequency signal from the synthetic image are combined to get the blended output.}
 \label{fig:blend}
\end{figure}

In addition to improving the inference performance, we also fine-tune the network model to further adapt it to our volumetric reconstruction context. To this end, we create a 3D human body dataset captured in our volumetric studio context, called HBR dataset. It consists of about 12000 training samples from 5 different human performances, and each sample includes a reference image, its ground truth depth map, a certain amount of neighboring source images and the camera parameters of all images. An overview of the generation of the training dataset is illustrated in \autoref{fig:blend}. We apply another 3D human body reconstruction pipeline \cite{worchel2020ernst} to generate textured meshes from images, and then render synthetic images and corresponding depth maps from the same viewpoints as real images using Blender \cite{Blender}. Note that to improve data efficiency, the real input images are cropped beforehand using the same method as mentioned above. Similar to the training data generation approach in the BlendedMVS dataset \cite{yao2020blendedmvs}, high-frequency signals from rendered (also known as synthetic) images and low-frequency information from real images are blended to create the images in our training dataset. The zoomed-in region in \autoref{fig:blend} shows a case of imperfectly reconstructed shape: the blended image retains a similar lighting condition from the real image, while being geometrically consistent with the synthetic image and rendered depth map. By combining visual cues from a synthetic image which reflects the underlying geometry of the corresponding depth map and the realistic environmental lighting from the real image, we get rid of the domain shift issue which arises when performing inference on real images using a deep model trained from synthetic images. Different from the input images of other workflows, which are already in standard RGB color space, our input images are derived from debayering and color space transformation including gamma correction from raw images in linear RGB space. In order to be robust to these effects, we add additional random gamma processing and Gaussian noise as data augmentation of the training data.

\section{Depth Map Filtering and Fusion} \label{DepthMapFusion}
Given dense depth maps for each view, we filter out noise and outliers and keep reliable depths before fusing them into global 3D space. Specifically, we apply four criteria for depth map filtering:
\begin{itemize}[leftmargin=*]
\item Photometric Confidence. An important advantage of cost volume based depth regression is that a confidence map can be explicitly derived from the probability distribution along depth direction. Based on the assumption of an unimodal probability distribution \cite{yao2018mvsnet}, we calculate the confidence map by summing over probabilities within a local window of the regressed depth estimation using softmax. As the confidence value is primarily a product from photometric feature matching, it is also called photometric consistency, ranging from 0 (inconsistent) to 1 (consistent). We filter out the depth values where the corresponding photometric confidence values are lower than a threshold denoted by $\tau_{photo}$.
\item Geometric Consistency. Similar to other MVS approaches \cite{schoenberger2016mvs,yao2018mvsnet,zhang2020visibilityaware}, we enforce depth observations to be geometrically consistent with neighboring views. Specifically, we project the reference depth map to neighboring views, and then reproject them back to the reference image through neighboring depth maps. Depth estimation of a pixel is considered geometrically consistent with another view if both the reprojection error in pixel coordinates is smaller than $\tau_{pix}$ and the relative deviation of reprojected depth is less than $\tau_{dep}$. In order to be robust against outliers, depths are retained only if they are geometrically consistent with at least $\tau_{geo}$ other views.
\item Foreground Mask. In contrast to general scene reconstruction, only foreground objects are of interest in volumetric reconstruction. To this end, a statistic background model is trained from clean plate images to facilitate segmentation for each camera. The usage of segmentation masks is twofold: only depth estimations belonging to the foreground region is fused into global 3D space; the extracted global point cloud is projected back to every camera, and points that fall into the background area of any camera view are filtered out.
\item Viewing Frustum. We notice that using the three criteria above still cannot remove all background outliers, because the foreground masks are not perfectly accurate and may include minor background pixels. Based on the observation that most but not always all cameras cover the entire scene of human performance, we filter out points which are outside the frustum boundary of at least $\tau_{frustum}$ cameras, which is similar to visible subvolume division \cite{worchel2020ernst}. In comparison, we employ the frustums of cropped cameras instead of the original cameras, resulting in enhanced efficiency.
\end{itemize}

Inspired by a recent volumetric reconstruction pipeline \cite{worchel2020ernst}, we leverage both the robustness against random noise of the TSDF volume \cite{curless1996volumetric} and the effective capability of water-tight mesh restoration of Screened Poisson surface reconstruction \cite{kazhdan2013screened} to integrate multi-view filtered depth maps into the global 3D space. Instead of directly back-projecting filtered depth maps into a global point cloud \cite{schreer2019capture,collet2015high,Relightables}, we build up a spatial TSDF volume, where depth information is fused. Using TSDF volumes not only alleviates random noise in the depth estimation, but can also be adaptive to the desired level of geometric details by adjusting the voxel resolution. Then, an oriented point cloud is extracted from the iso-surface of the volume before meshing by Screened Poisson surface reconstruction. Comparisons for the different filtering and fusion techniques are presented in \autoref{Ex_FU}.

\section{Experiments} \label{Experiment}

\subsection{Implementation}
We evaluate our work on a variety of volumetric human performance data from two main perspectives: depth map estimation and post-processing. In the data pre-processing stage, we crop an adaptive ROI from the original images of resolution $5120 \times 3840$. In our experiments, depending on the size of human appeared in the input image, the ROI height ranges from 2500 to 3800 while the width ranges from 1200 to 1700, while the depth sampling number $D$ is set at 288. For MVS depth estimation, we first train a Vis-MVSNet model on the DTU dataset \cite{jensen2014large} using recommended hyperparameters \cite{zhang2020visibilityaware}. For both fine-tuning and testing on human body datasets, we select 5 neighbouring source images for each reference image based on the viewing angle criterion \cite{yao2018mvsnet}. Moreover, we optimize and fine-tune the networks to better adapt to our human reconstruction context, as discussed in \autoref{DepthMapEstimation}. The estimated depth map has half the size of the input ROI image. Regarding depth map filtering and fusion, we set the threshold values $\{\tau_{photo},\tau_{pix},\tau_{dep},\tau_{geo},\tau_{frustum}\} = \{0.8, 1, 0.01, 2, 2\}$ to discriminate outliers, while the resolution of TSDF volume is set to 2 mm. Our experiments are implemented on a single machine with Intel Xeon E5-2643 CPU and one NVIDIA TITAN RTX GPU. The processing time of our full volumetric reconstruction pipeline (from input images to streamable MP4 files) is around 5 min/frame on the single machine. It is worth noting that most stages (except mesh registration) can be parallelized by distributing data per camera or per frame on multiple GPUs or machines.
% The GPU memory bottleneck lies in MVS depth estimation and depends on the size of the cropped ROI, as described in \autoref{table:memory_DE}. 

% \newcolumntype{Y}{>{\centering\arraybackslash}X}
% \def\tabularxcolumn#1{m{#1}}
% \begin{table}[t]
% \centering
% \begin{tabularx}{\columnwidth}{Y|Y|Y}
%  \hline
% Input resolution & GPU Mem. (MB) & Runtime (s) \\ 
%  \hline
% 2500$\times$1200 & 6841 & 3.59 \\
% 3800$\times$1700 & 13715 & 6.24 \\
%  \hline
% \end{tabularx}
% \caption{Comparison of GPU memory footprint and runtime with different sizes of cropped ROI inputs for MVS depth estimation.}
% \label{table:memory_DE}
% \end{table}

\begin{figure*}[t]
 \centering % avoid the use of \begin{center}...\end{center} and use \centering instead (more compact)
 \includegraphics[width=0.85\textwidth]{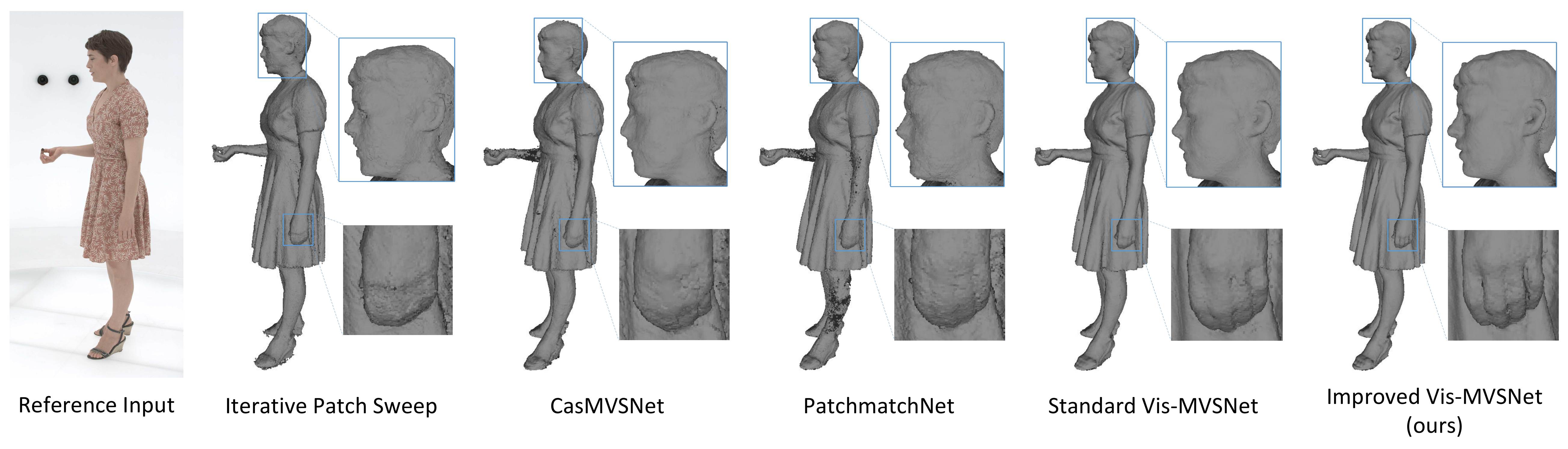}
 \caption{Comparison of different depth estimation methods. The leftmost image is a sample from 32 cropped input images as reference, while others are resulting fused oriented point clouds using different depth estimation approaches.}
 \label{fig:DE}
\end{figure*}

\def\tabularxcolumn#1{m{#1}}
\begin{table}[t]
\centering
\begin{tabularx}{\columnwidth}{XXXXXX}
 \hline
 & Iterative Patch Sweep \cite{worchel2020ernst} & Cas-MVSNet \cite{gu2020cascade} & Patch-Match-Net \cite{wang2020patchmatchnet} & Standard Vis-MVSNet \cite{zhang2020visibilityaware} & Ours \\ 
 \hline
IoU(\%) & 96.59 & 95.42 & 95.95 & 98.01 & \textbf{98.88} \\
 \hline
\end{tabularx}
\caption{Average IoU of ground truth and rendered silhouettes of different depth estimation methods.}
\label{table:DE}
\end{table}

\subsection{Comparisons in Depth Estimation} \label{Ex_DE}
We first evaluate the reconstructed 3D geometry using different depth estimation methods including stereo matching and MVS. \autoref{fig:DE} compares a traditional PatchMatch based stereo matching approach called Iterative Patch Sweep \cite{Waizenegger2016,worchel2020ernst}, deep MVS based architectures including CasMVSNet \cite{gu2020cascade}, PatchmatchNet \cite{wang2020patchmatchnet} and Vis-MVSNet \cite{zhang2020visibilityaware}, as well as our optimized and tuned Vis-MVSNet. Note that all learning-based MVS networks are pre-trained on the DTU dataset \cite{jensen2014large}, and we use the same proposed depth map post-processing approach for fair comparison. While both CasMVSNet and PatchmatchNet fail in homogeneous skin areas such as the arm and leg, Iterative Patch Sweep and standard Vis-MVSNet are more robust against low-textured regions. With visibility dependent multi-level cost volume regression, standard Vis-MVSNet outperforms other pre-trained networks by having relatively higher geometric accuracy, as shown in zoomed in head and hand regions. However, it is still not sufficient for many applications where fine details are desired. Benefiting from proposed optimizations and tuning in volumetric capturing context, our improved Vis-MVSNet achieve significantly more accurate geometry reconstruction with high-resolution details, e.g., clearly distinguishable fingers.

\begin{figure*}[t]
 \centering % avoid the use of \begin{center}...\end{center} and use \centering instead (more compact)
 \includegraphics[width=0.8\textwidth]{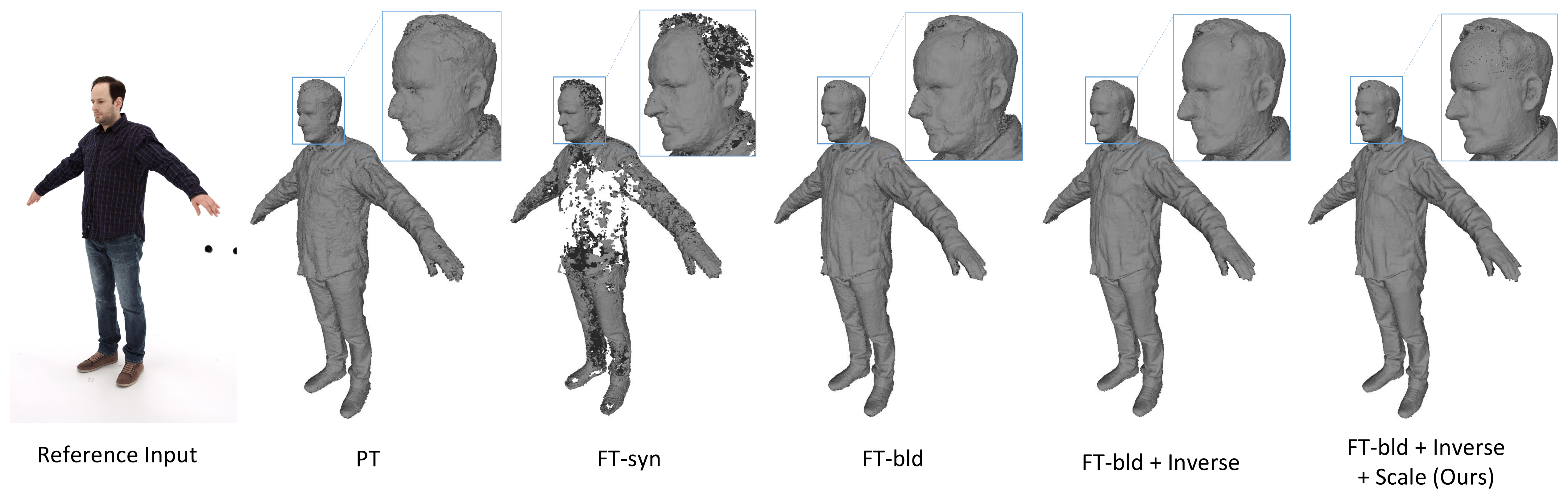}
 \caption{Ablation experiments with different configurations of Vis-MVSNet for depth estimation on an unseen human. To differentiate the results, one can pay attention to the lip, nose and ear of the human.}
 \label{fig:vismvsnet}
\end{figure*}

\def\tabularxcolumn#1{m{#1}}
\begin{table}[t]
\centering
\begin{tabularx}{\columnwidth}{XXXXXX}
 \hline
 & PT & FT-syn & FT-bld & FT-bld +Inv. & FT-bld +Inv.+Scl. (Ours) \\ 
 \hline
IoU(\%) & 98.01 & 63.92 & 98.21 & 98.54 & \textbf{98.88} \\
 \hline
\end{tabularx}
\caption{Average IoU of ground truth and rendered silhouettes of different configurations of Vis-MVSNet. PT: official pre-trained model, FT-syn: fined-tuned with synthetic images of HBR dataset, FT-bld: fined-tuned with blended images, Inv.: sampling in inverse depth space, Scl.: aggressively scaled down depth search range between consecutive stages.}
\label{table:vismvsnet}
\end{table}

For quantitative evaluation, we consider the foreground segmentation mask from clean plate subtraction as the ground truth silhouette, to circumvent the limitation of the lack of 3D ground truth data. Specifically, we reconstruct meshes using various depth estimation methods and render them into the viewpoints of input cameras. To compare the similarity between rendered mask and ground truth silhouette, we adopt Intersection over Union (IoU). We compute the IoU for each method averaging over all frames and all cameras. As shown in \autoref{table:DE}, our optimized Vis-MVSNet has the highest IoU, which indicates superior geometric accuracy compared to other methods.
%  Specifically, the proposed method outperforms the traditional Iterative Patch Sweep by more than 2\% in the IoU metric.

To further evaluate the individual contributions of our optimizations based on the standard Vis-MVSNet, we perform an ablation study on different configurations of the neural network on a human which is not in the training set. As illustrated in \autoref{table:vismvsnet} and \autoref{fig:vismvsnet}, fine-tuning the pre-trained Vis-MVSNet model using blended images from our 3D human body dataset described in \autoref{DepthMapEstimation} improves the network performance, while using synthetic images deteriorate it significantly as it overfits to the synthetic color domain. This demonstrates the importance of blending real and synthetic images to alleviate the domain gap. Moreover, both sampling depth candidates in inverse space and scaling down more aggressively the depth search range from coarse to fine stages lead to finer and more accurate details in the reconstructed human models (see lip, nose and ear in \autoref{fig:vismvsnet}), as well as better quantitative results (ours outperforms fine-tuned standard Vis-MVSNet by 0.67\% in the IoU metric). In addition, the high-fidelity geometric details on the unseen data demonstrate strong generalization ability of the networks.

\begin{figure}[t]
     \centering
     \begin{subfigure}[b]{0.168\columnwidth}
         \centering
         \includegraphics[width=\textwidth]{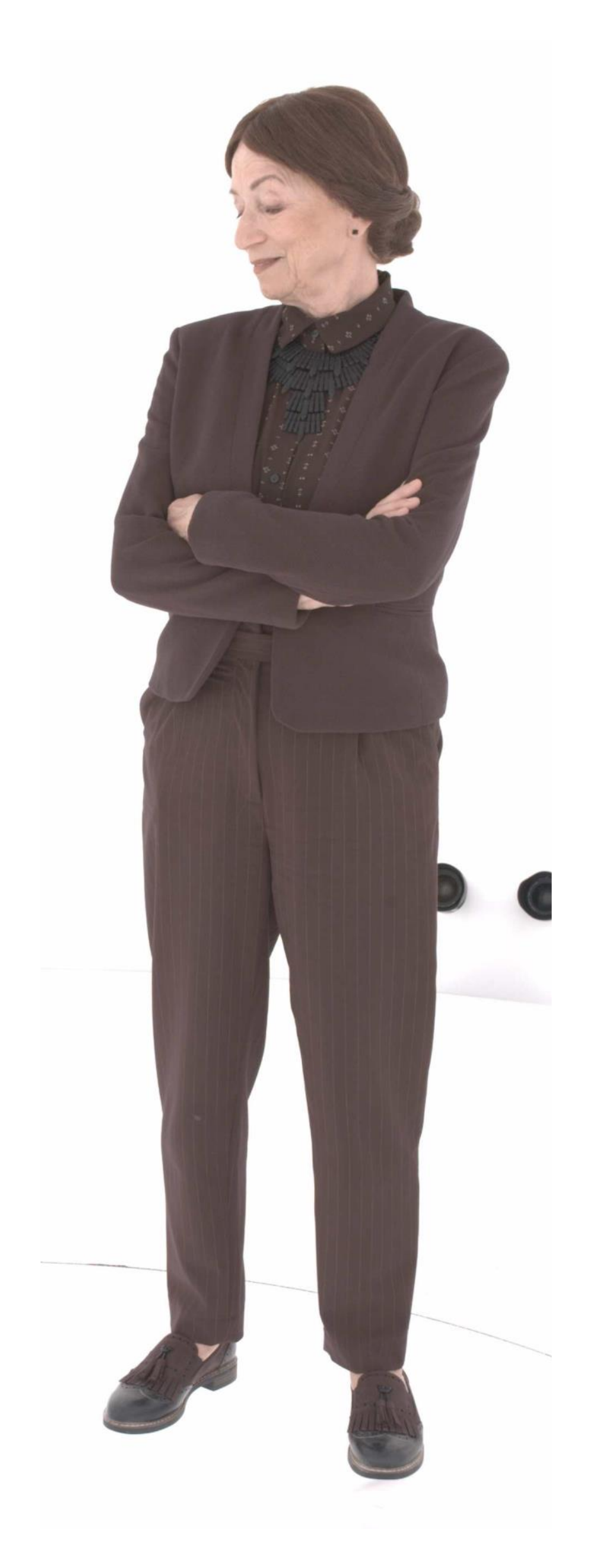}
        %  \caption{Input image example}
         \caption{}
         \label{fig:fu1}
     \end{subfigure}
    %  \hfill
     \begin{subfigure}[b]{0.19\columnwidth}
         \centering
         \includegraphics[width=\textwidth]{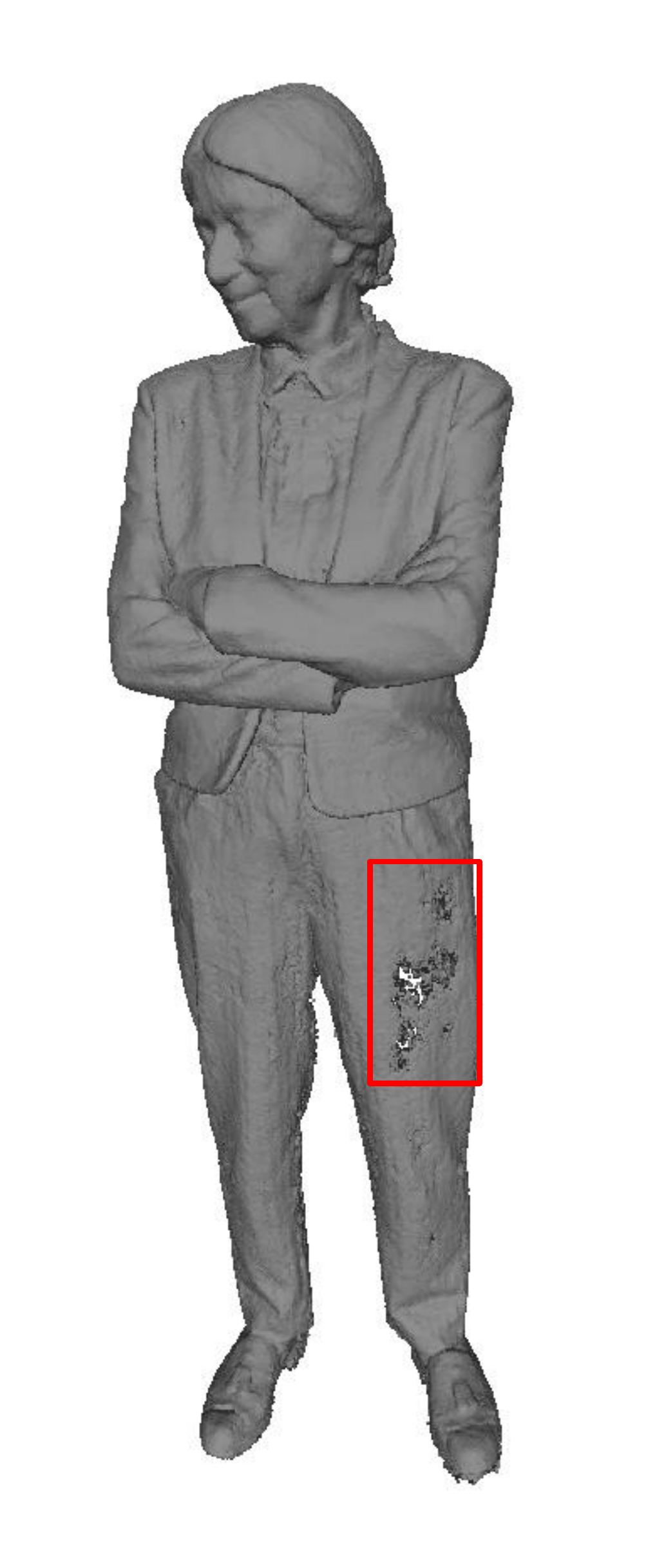}
        %  \caption{Without PC and GC \newline}
        %  \caption{W/o PC and GC}
         \caption{}
         \label{fig:fu2}
     \end{subfigure}
    %  \hfill
     \begin{subfigure}[b]{0.19\columnwidth}
         \centering
         \includegraphics[width=\textwidth]{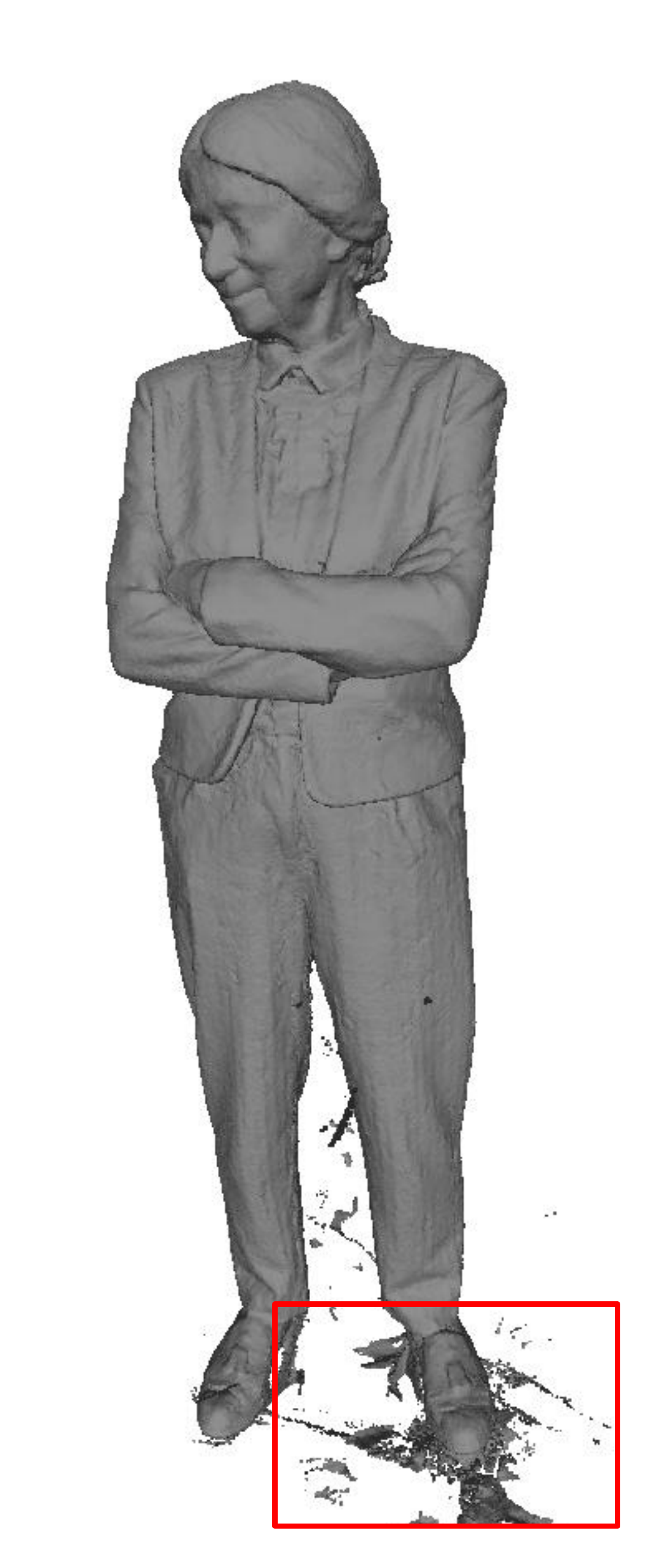}
        %  \caption{Without FM and VF \newline}
        %  \caption{W/o FM and VF}
         \caption{}
         \label{fig:fu3}
     \end{subfigure}
    %  \hfill
     \begin{subfigure}[b]{0.19\columnwidth}
         \centering
         \includegraphics[width=\textwidth]{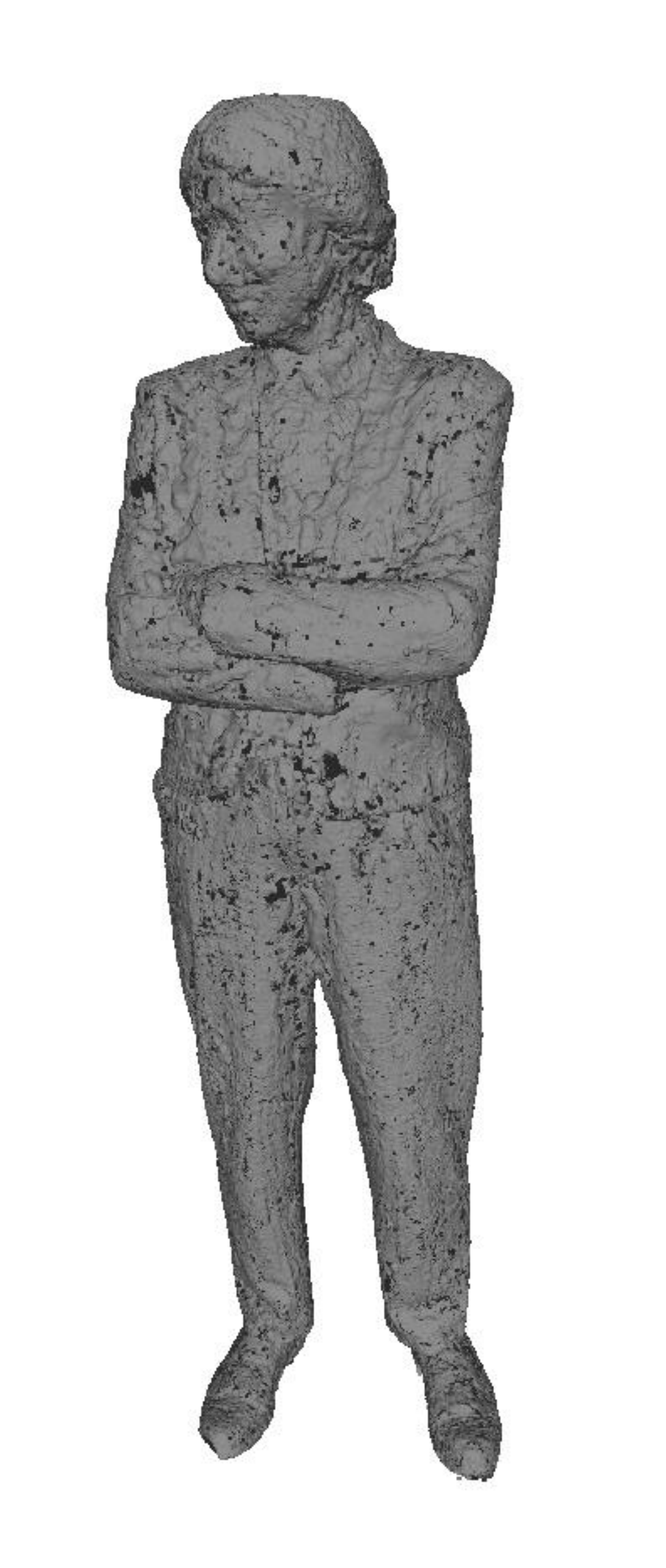}
        %  \caption{Point based}
         \caption{}
         \label{fig:fu4}
     \end{subfigure}
    %  \hfill
     \begin{subfigure}[b]{0.19\columnwidth}
         \centering
         \includegraphics[width=\textwidth]{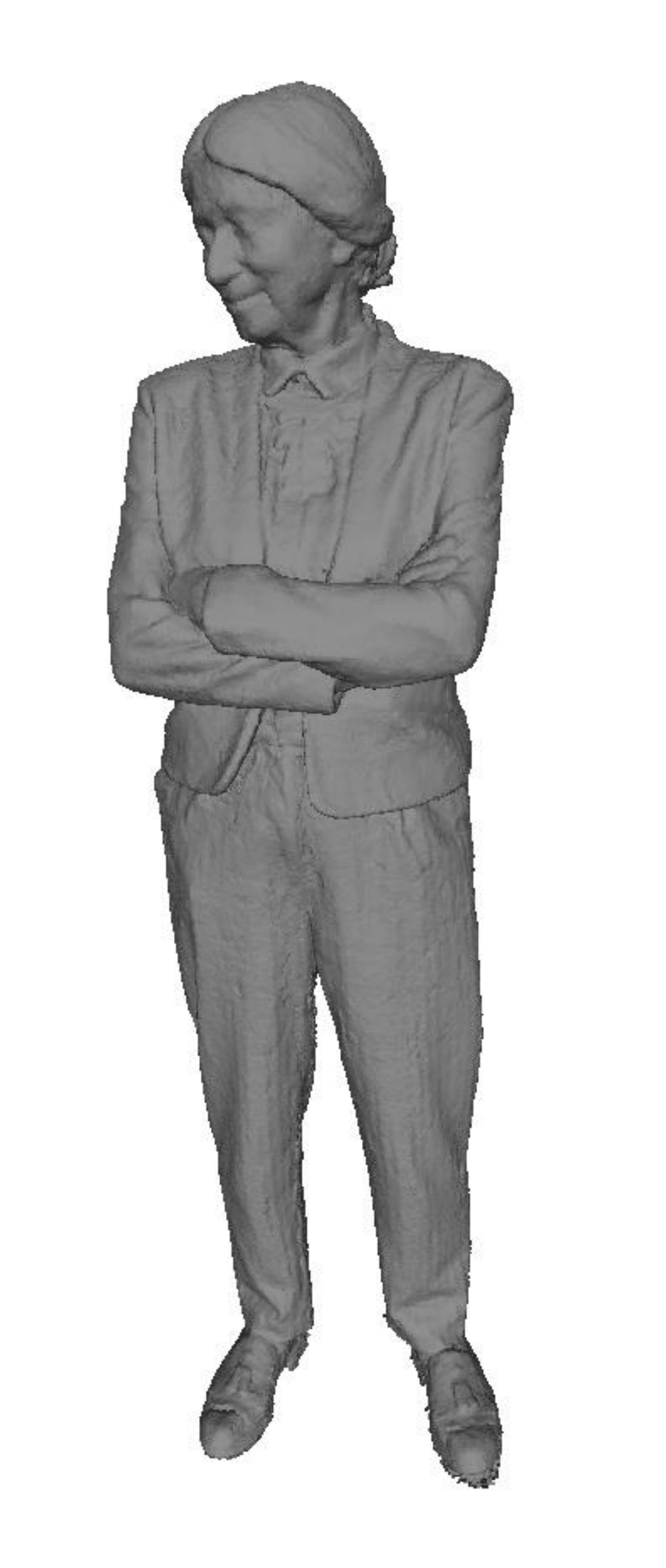}
        %  \caption{Ours \newline}
        %  \caption{Ours (PC+GC+FM+VF)}
         \caption{}
         \label{fig:fu5}
     \end{subfigure}
        \caption{Comparison of fused oriented point clouds using different depth map post-processing configurations. (a): Input image example, (b): W/o PC and GC, (c): W/o FM and VF, (d): Point based fusion, (e): Ours (TSDF fusion)}
        \label{fig:FU}
\end{figure}
% Ours (PC+GC+FM+VF)

\begin{table}[t]
\centering
\begin{tabularx}{\columnwidth}{XXXXX}
 \hline
 & Without PC\&GC & Without FM\&VF & Point-based fusion & Ours (TSDF fusion) \\ 
 \hline
IoU(\%) & 95.51 & 96.60 & 98.22 & \textbf{98.88} \\
 \hline
\end{tabularx}
\caption{Average IoU of ground truth and rendered silhouettes of different depth post-processing configurations.}
%  & Without PC\&GC & Without FM\&VF & Point-based fusion & PC+GC +FM+VF (Ours) \\ 
% PC: Photometric Confidence, GC: Geometric Consistency, FM: Foreground Mask, VF: Viewing Frustum
\label{table:FU}
\end{table}

\subsection{Comparisons in Depth Post-Processing} \label{Ex_FU}
We qualitatively evaluate the effectiveness of our proposed depth map post-processing method consisting of filtering and fusion. In this work, four criteria for depth map filtering are implemented: Photometric Confidence (PC), Geometric Consistency (GC), Foreground Mask (FM), Viewing Frustum (VF), while we also compare voxel-based and point-based fusion. Details of these principles are discussed in \autoref{DepthMapFusion}. \autoref{fig:FU} illustrates fused oriented point clouds from different combinations of depth filter and fusion, using the same estimated depth maps as input. Our proposed approach considers all of the four filtering criteria and applies voxel-based TSDF volume fusion, as shown in \autoref{fig:fu5}, and serves as reference for comparing other configurations. By removing photometric confidence and geometric consistency filtering (\autoref{fig:fu2}), inconsistent and error-prone depth estimations, which are often presented in areas with low textureness or repetitive patterns (e.g., pants), are also integrated into global geometry and thus interfere with other correct, high-confidence estimations. On the other hand, disabling depth filtering from using the foreground mask and viewing frustum brings noisy and undesired points from the background (see \autoref{fig:fu3}). Note that the depth estimation in the background is usually incorrect because (1) the predefined depth search range is calculated only for foreground objects and (2) background areas suffer from higher occlusion and less common visibility from multiple views. Moreover, we also create a result using the same proposed filtering technique but directly fused into a point cloud, as shown in \autoref{fig:fu4}, which suffers from holes and noise and requires extra efforts on decimation and smoothing before meshing. In contrast, our proposed method (see \autoref{fig:fu5}) benefits from TSDF volume, which is capable to smooth out noise from depth maps and output 3D geometry of desired level of resolution. We again provide an evaluation on the IoU between rendered and ground truth silhouette, as illustrated in \autoref{table:FU}. Similar to our qualitative observation, our method performs the best in quantitative comparison.

% \subsection{VR use case}
% Benefiting from highly accurate geometry reconstruction using the proposed method, along with efficient temporal mesh registration \cite{morgenstern2019progressive} and data stream compression, we achieve high-fidelity human performance reconstruction for volumetric video, as presented in the supplementary video. Our reconstruction result can be integrated into a variety of applications, such as VR, as demonstrated in \autoref{fig:arvr} where the reconstructed human acts as a speaker on the stage.

% \begin{figure}[tb]
%  \centering % avoid the use of \begin{center}...\end{center} and use \centering instead (more compact)
%  \includegraphics[width=\columnwidth]{hanna_all.png}
%  \caption{A VR application from our work uses a basic CG environments to underline the realistic and natural impression of real person based volumetric video assets.}
%  \label{fig:arvr}
% \end{figure}

\section{Conclusion}
We present a novel professional volumetric video pipeline for human body reconstruction using only 32 passive cameras. We introduce and optimize learning-based MVS networks to estimate depth maps for highly geometrically accurate human body reconstruction, and further adapt them to the ultra-high resolution video input. Furthermore, we propose a novel depth map post-processing approach consisting of filtering by comprehensive criteria and fusion into TSDF volume, to create a clean and coherent point cloud. Experiments demonstrate qualitatively and quantitatively effectiveness of the proposed method. In spite of promising performance on general human body reconstruction, our method suffers from recovering reflective objects like glasses, because non-Lambertian surfaces pose a challenge for illumination-agnostic MVS. An interesting future work direction is to use view-dependent scene representation like implicit neural radiance field, which gains considerable popularity recently.

% We have presented our improved end-to-end human body reconstruction pipeline to produce high-quality volumetric video using only 32 passive cameras. We optimize and tune state-of-the-art learning-based MVS networks for highly geometrically accurate human body reconstruction. Furthermore, we propose a novel depth map post-processing approach consisting of filtering by comprehensive criteria and fusion into TSDF volume, to create a clean and coherent point cloud. Extensive experiments demonstrate effectiveness of the proposed method and its impressive application.

\noindent
\textbf{Acknowledgements}  This work was partially supported by the Investitionsbank Berlin in the ProFIT funding scheme with financial support of European Regional Development Fund (EFRE) and the government of Berlin in the context of the KIVI project and the European Union’s Horizon 2020 research and innovation program under Grant Agreement 952147 (Invictus). We also thank INVR.SPACE GmbH and Volucap GmbH for their support.

% BibTeX files (here: refs). 
% -------------------------------------------------------------------------
% \bibliographystyle{plain}
\bibliographystyle{abbrv}
\bibliography{refs}

\end{document}